\documentclass[letterpaper, 10 pt, conference]{./ieeeconf/ieeeconf}  % Comment this line out if you need a4paper

\IEEEoverridecommandlockouts                              % This command is only needed if 
\usepackage{epsfig} % for postscript graphics files
\usepackage{times} % assumes new font selection scheme installed
\usepackage{amsmath} % assumes amsmath package installed
\usepackage{amssymb}  % assumes amsmath package installed
\usepackage{xcolor}
\usepackage{epsfig} % for postscript graphics files
\usepackage{bytefield}
\usepackage{bm}
\usepackage[binary-units = true]{siunitx}
\usepackage[normalem]{ulem} %Makes \emph{} underline text instead of italics....

\usepackage{balance}
\usepackage{graphicx}
\usepackage{nth}
\usepackage[hidelinks]{hyperref}
\usepackage{caption} % for proper captioning of our tables
\newcommand{\SB}{SUPERball} %use \SB{} to write "SUPERball" correctly

\title{\LARGE \bf
State Estimation for Tensegrity Robots 
}

\author{Ken Caluwaerts$^{1,*}$, Jonathan Bruce$^{2,*}$, Jeffrey M. Friesen$^{3}$, and Vytas SunSpiral$^4$
\thanks{*These authors contributed equally to this work.}% <-this % stops a space
\thanks{This work was performed while all authors were at the \mbox{NASA Ames Research Center}, Moffett Field CA, USA, with funding from NASAs NSTRF, NIAC and GCD Programs.}
\thanks{$^{1}$Oak Ridge Associated Universities (ORAU),
        Oak Ridge TN, USA
        {\tt\small ken.caluwaerts@nasa.gov}}%
\thanks{$^{2}$University of California Santa Cruz,
        Santa Cruz CA, USA
        {\tt\small jebruce@ucsc.edu}}%
\thanks{$^{3}$University of California San Diego,
        San Diego CA, USA
        {\tt\small jfriesen@ucsd.edu}}%
\thanks{$^4$SGT Inc., Greenbelt MD, USA
	{\tt\small vytas.sunspiral@nasa.gov}}
}

\begin{document}

\maketitle
\thispagestyle{empty}
\pagestyle{empty}

%%%%%%%%%%%%%%%%%%%%%%%%%%%%%%%%%%%%%%%%%%%%%%%%%%%%%%%%%%%%%%%%%%%%%%%%%%%%%%%%
\begin{abstract}

Tensegrity robots are a class of compliant robots that have many desirable traits when designing mass efficient systems that must interact with uncertain environments. 
Various promising control approaches have been proposed for tensegrity systems in simulation.
Unfortunately, state estimation methods for tensegrity robots have not yet been thoroughly studied. 
In this paper, we present the design and evaluation of a state estimator for tensegrity robots.
This state estimator will enable existing and future control algorithms to transfer from simulation to hardware.
Our approach is based on the unscented Kalman filter (UKF) and combines inertial measurements, ultra wideband time-of-flight ranging measurements, and actuator state information.
We evaluate the effectiveness of our method on the \SB{}, a tensegrity based planetary exploration robotic prototype. 
In particular, we conduct tests for evaluating both the robot's success in estimating global position in relation to fixed ranging base stations during rolling maneuvers 
as well as local behavior due to small-amplitude deformations induced by cable actuation.

\end{abstract}

\section{Introduction}
\label{intro}

Tensegrity robotics is a relatively young field of research wherein a robot is structured according to tensegrity principles. 
We define a tensegrity as a structure with discontinuous compression elements suspended within a web of tension elements. 
In this class of robots, motion is often achieved through actuation of the tensile elements within the structure. 
%High structural mass efficiency, system-wide compliance, re-distribution of external forces through the tension network, and the ability to pack into a tight space and self deployment are some of the benefits such robots promise to deliver. 

Tensegrity robots have highly coupled dynamics due to their interconnected network of compliant tensile elements. 
As such, an external force exerted at a single point will cause a global displacement in all nodal positions of the system. 
This property is beneficial in that it allows the system to passively adapt to external forces and redistribute loads effectively through the tension network.
However, it also causes difficulties in determining the state of the system when only limited sensor information is available. 

Various control approaches have been proposed for tensegrity systems in simulation.
However, these algorithms often rely on full state or trajectory information~\cite{Rieffel2007}\cite{Paul2005}\cite{GraellsRovira2009}\cite{sultan2000tensegrity}.
In this paper, we present the design and evaluation of a state estimator for this class of robot
which will allow the transfer of these existing and future control algorithms from simulation to hardware.

\begin{figure}[tpb]
 \centering
  \includegraphics[width=.9\linewidth]{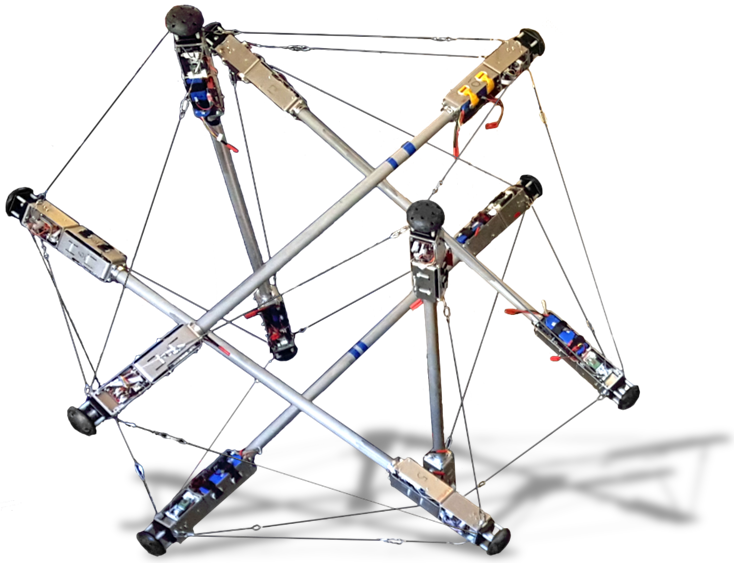}
 \caption{The \SB{} prototype, a six strut tensegrity robot with 12 actuated cables and 12 passive cables. Our state estimation algorithm is evaluated on this platform.}
\label{fig:SUPERball_robot}
 \end{figure}

We focus on the use of low-cost ranging modules and inertial measurement units mounted to the rods of a tensegrity robot as the sensor inputs to an unscented Kalman filter (UKF).
%To estimate the robot's relative global position, eight more low-cost ranging modules are placed external to the robot. 
%Eight ranging base stations, external to the robot, are utlized which allow for the robot's position relative to these stations to be estimated. 
% To improve accuracy, the sensor information from these base stations is combined with data from inertial measurement units (IMUs) located on the robot and motor encoders.  
These ranging sensors can be purchased off-the-shelf and do not rely on any user-designed mechanical infrastructure to operate. 
We also use motor encoders to sense change in cable rest length as control inputs %fed 
into the dynamic model utilized by the UKF.
For testing our approach, we use the \SB{} prototype, a six strut tensegrity robot designed to explore tensegrity systems for planetary exploration~\cite{bruce2014design}\cite{sabelhaus2015system}. 
% The following sentence is out of scope of this paper and kind of out of context to the rest of the intro.
%Its design is radically different from existing wheeled planetary rovers, allowing it to better address some challenges faced by these robots.
\SB{} is shown in Fig.~\ref{fig:SUPERball_robot}. 

% We utilize eight ranging base stations which allow for the robot's position relative to these stations to be estimated. 
% To improve accuracy, the sensor information from these base stations is combined with data from inertial measurement units (IMUs) located on the robot and motor encoders. 
%These external base stations are not required for determining the local state of the robot, 
%but with the minimal sensor information currently fed into the filter the additional measures provided by these base stations improves accuracy substantially. 

This paper is organized as follows.
We first present a detailed overview of the system's sensors, ranging method and calibration routine. 
We then describe the implementation of the unscented Kalman filter and the dynamics model. 
This is followed by our experimental results. 
We end this paper with our conclusions and future outlook.

%%%%%%%%%%%%%%%%%%%%%%%%%%%%%%%%%%%%%%%%%%%%%%%%%%%%%%%%%%%%%%%%%%%%%%%%%%%%%%%%
\section{\SB{} Overview}
\label{overview}

The Spherical Underactuated Planetary Exploration Robot (\SB{}) is a preliminary tensegrity robot prototype developed at the NASA Ames Research Center. 
The purpose the \SB{} project is to develop technologies for a new class of planetary exploration robot which is able to 
deploy from a compact launch volume, land at high speeds without the use of air-bags, and provide robust surface mobility.  This broad set of functions can be enabled by utilizing the efficiency and structural compliance of tensegrity robots.  Passive-structure drop tests have confirmed the analysis supporting the high-speed landing concept, and the current prototype of \SB{} is intended to develop the foundational engineering approaches required to support surface locomotion by tensegrity robots. 
A full system overview is out of the scope of this paper, but the relevant details of the system will be discussed.
Please refer to Bruce et. al.~\cite{bruce2014design} and Sabelhaus and Bruce~\cite{sabelhaus2015system} for system details.

Relevant to this work are the sensors, actuators, and the ROS network implemented on \SB{}.
\SB{} consists of 6 identical rods held together by 24 cables in-line with springs.
As described in~\cite{bruce2014design}, each rod of \SB{} is comprised of two modular tensegrity robotic platforms (end caps).
Each platform is equipped with inertial measurement units, a motor with encoders, and a fully enabled Robot Operating System (ROS) node which communicates to our ROS network via WiFi.
\SB{} is underactuated and only 12 out of the 24 cables can be actuated (shortened) by spooling cable around a spindle (one per end cap).
This allows the robot to roll through deformation.
Since each modular tensegrity platform can be outfitted with sensors, ranging sensors were added for this paper to enable positioning.
The ranging sensors are discussed in detail in Section~\ref{ranging}.

%%%%%%%%%%%%%%%%%%%%%%%%%%%%%%%%%%%%%%%%%%%%%%%%%%%%%%%%%%%%%%%%%%%%%%%%%%%%%%%%
\section{Ranging Setup and Calibration}
\label{ranging}
\label{txt:ranging}
This section introduces the hardware and software setup for a set of wireless ranging modules to enable 
position tracking of the robot both as an as internal distance measurements (end cap to end cap) an in an external (world) reference frame.

We equipped all end caps of \SB{} with a DWM1000 ranging module from DecaWave Ltd.
By employing ultra wideband technology, the low-cost DWM1000 modules provide wireless data transfer and highly accurate timestamps of transmitted and received packets. 
This allows the distance between two DWM1000 modules to be estimated by computing the time-of-flight of exchanged messages without the need for synchronized clocks.
We opted for this technology because it allows proprioceptive state estimation (distances between end caps), which 
cannot be easily tracked directly via motor encoders.~\cite{ledergerber2015}   Furthermore, we 
placed eight more DWM1000 modules as "fixed anchors" around our testing area to provide a world reference frame for
 ground truth and generation of a reward signal for the machine learning algorithms that we will use to develop 
 locomotion controllers for the robot.  Our intention is that the fixed anchors will not be required in the final deployed 
 version of the robot, and are primarily for use during algorithm development.

We first introduce the basic sensor operation and our approach to efficiently estimate distances between a large number of sensor modules.
This is followed by a discussion of our ranging software and hardware setup.
Finally, we provide a calibration routine similar to a common motion capture system that allows for quick set up of the sensor network.

\subsection{Sensor Operation}
\subsubsection{Bidirectional Ranging}
We operate the DWM1000 in the so-called \emph{symmetric double-sided two-way ranging} mode.
In this mode, the modules exchange $3$ packets to estimate the time-of-flight between each other.
While the time-of-flight of unsynchronized modules can be estimated with the exchange of only $2$ packets, the employed mode can significantly reduce measurement noise~\cite{decawave}.

\begin{figure}[tpbh]
 \centering
  \includegraphics[width=.6\linewidth]{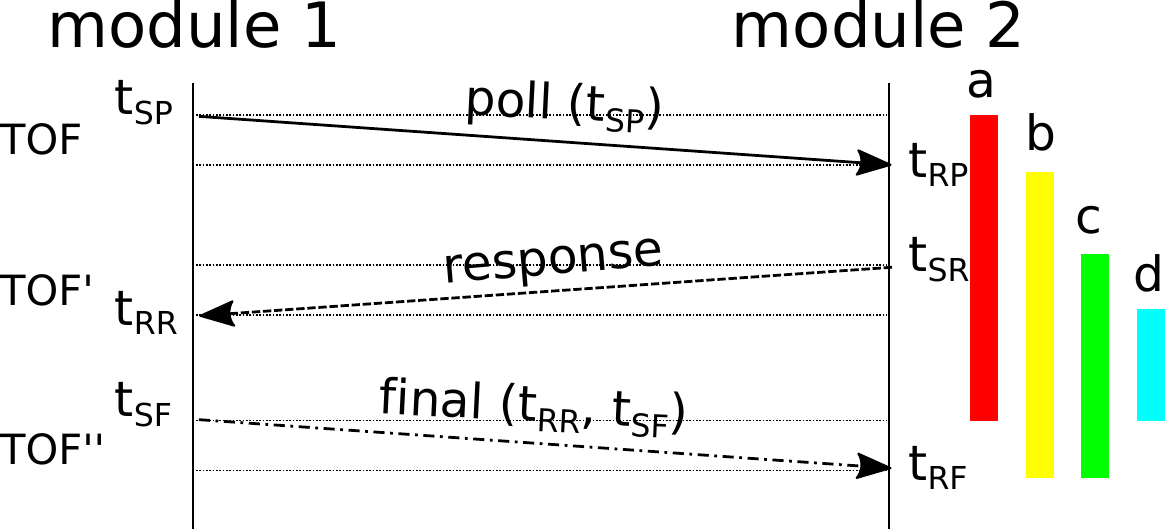}
 \caption{Basic symmetric double-side two-way ranging packet exchange. 
 Modules 1 and 2 exchange 3 packets (\emph{poll}, \emph{response}, and \emph{final}). Module 2 then estimates the distance between the modules based on the local timestamps.}
\label{fig:bidirectional_ranging}
 \end{figure}

The basic ranging packet exchange is shown in Fig.~\ref{fig:bidirectional_ranging}.
One module sends out a \emph{poll} message containing an emission timestamp ($t_{SP}$) using its local clock.
A second module receives this message and timestamps the time of arrival using its local clock ($t_{RP}$).
The second module then sends out a \emph{response} packet at time $t_{SR}$ (module 2's clock).
The first module receives this packet at time $t_{RR}$  (module 1's clock).
Module 1 now sends out a final message containing $t_{RR}$ and the emission time of the final message ($t_{SF}$, clock of module 1).
Module 2 receives this information and timestamps it ($t_{RF}$).

Module 2 can now estimate the time-of-flight and the distance between itself and module 1 based on the 6 timestampes.
The basic equations to estimate the distance between module $i$ and module $j$ (module $i$ initiates the ranging and module $j$ computes the distance) are given by:
\begin{eqnarray} %ranging between i and j (j computes distance)
a_{i} &=& t_{SF}^i-t_{SP}^i\\
b_{j,i} &=& t_{RF}^{j,i}-t_{RP}^{j,i}\\ %when j receives i
c_{j,i} &=& t_{RF}^{j,i}-t_{SR}^j\\  
d_{i,j} &=& t_{SF}^i-t_{RR}^{i,j} %i receives from j
\end{eqnarray}
\begin{eqnarray}
{TOF}_{j,i}  &\approx& \frac{1}{2}\left(c_{j,i}-d_{i,j}\frac{b_{j,i}}{a_i} \right)-\delta_{j,i}\\
\|\bm{N}_j - \bm{N}_i\| &\approx& \frac{1}{2\bm{c}}\left(c_{j,i}-d_{i,j}\frac{b_{j,i}}{a_i} \right)-o_{j,i} \label{eq:distance_estimation}\\
&\doteq& m_{j,i}-o_{j,i} .
\end{eqnarray}
The variables $a$, $b$, $c$, and $d$ are also visualized in Fig.~\ref{fig:bidirectional_ranging}.
The time-of-flight calculation between two modules $i$ and $j$ ($TOF_{j,i}=TOF_{i,j}$) is hindered by a fixed measurement offset ($\delta_{j,i}=\delta_{i,j}$).
This offset is due to antenna delays and other discrepancies  between the timestamps and actual packet reception or emission.
Whereas this offset is expected to be unique to each module, we found that it is necessary to estimate this offset pairwise for closely located modules.
Our hypothesis is that the proximity of the robot's motors and the sensor's position near the end cap's metal structure influence the antenna characteristics between pairs of modules.

Eq.~\ref{eq:distance_estimation} estimates the distances between the modules based on the time-of-flight calculation ($\bm{c}$ is the speed of light).
We rewrite the time offset $\delta_{j,i}$ as a distance offset $o_{j,i}$ (with $o_{j,i}=o_{i,j}$).
Here $\bm{N}_i$ and $\bm{N}_j$ refer to the positions of nodes $i$ and $j$ respectively (see Section~\ref{txt:ukf}).
The variables $m_{j,i}$ represent the uncorrected distance estimates.

%say offset symmetric

%lessons learned: antenna effect, bidirectional offset, restart, rx rx, 1ns pulses

The DWM1000 requires careful configuration for optimal performance.
We provide our main configuration settings in Table~\ref{tbl:dwm1000}.
The ranging modules tend to measure  non line-of-sight paths near reflective surfaces (e.g. floor, computer monitors), which may cause filter instability.
Using the DWM1000's built-in signal power estimator, we reject such suspicious packets. 
In practice, between $30\%$ and $70\%$ of packets are rejected in our indoor test environment.

\begin{table}[h]
\centering
\caption{DWM1000 configuration}
\label{tbl:dwm1000}
\begin{tabular}{llllll}
{\bf bitrate} & {\bf channel} & {\bf preamble} &  {\bf PRF} & {\bf preamble code} \\ \hline
\SI{6.8}{\mega\bit\per\second}     & 7             & 256                    & \SI{64}{\mega\hertz}     & 17                 
\end{tabular}
\end{table}

\subsubsection{Broadcast Ranging}
\begin{figure}[tpbh]
 \centering
  \includegraphics[width=.8\linewidth]{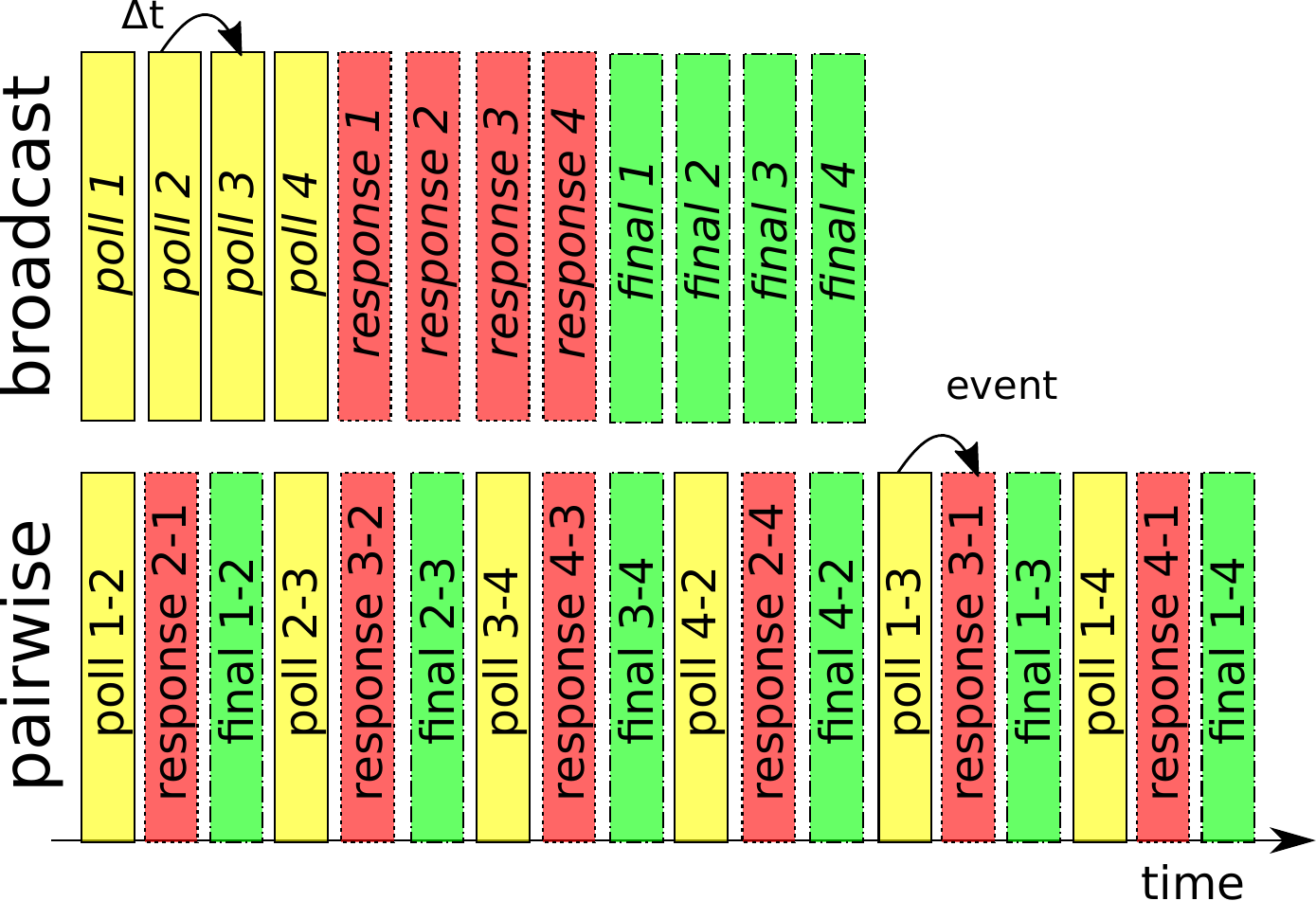}
 \caption{Packet exchange between 4 modules for bidirectional pairwise and broadcast ranging. 
 Timed broadcast messages allow for efficient ranging with a large number of modules.
  }
\label{fig:broadcast_ranging}
 \end{figure}

Due to the large number of exchanged packets (3 per pair) bidirectional ranging between pairs of modules quickly becomes inefficient when the number of modules grows.
We propose a simple approach using timed broadcast messages that scales linearly in the number of modules (3 packets per module).
In this setup one module periodically initiates a measurement sequence by sending out a \emph{poll} message.
When another module receives this message it emits its own \emph{poll} message after a fixed delay based on its ID, followed by \emph{response} and \emph{final} messages after additional delays.
Broadcast ranging is illustrated in Fig.~\ref{fig:broadcast_ranging}. 

One disadvantage of the broadcasting approach is that the total measurement time between a pair of modules takes longer (up to \SI{60}{\milli\second} in our experimental setup)
 than a single pairwise bidirectional measurement (approx. \SI{3}{\milli\second}).
However, broadcast ranging provides two measurements for each pair of modules per measurement iteration.

Note that each module now needs to keep track of the \emph{poll} and \emph{final} packet reception times of all other modules.
The \emph{final} packet becomes longer as each module needs to transmit the \emph{response} reception time ($t_{RR}$)  of all other modules.
%The simplified packet structures are:

{%\small
%\begin{bytefield}{22}
%  \begin{rightwordgroup}{poll}{
%    \bitbox{7}{preamble} &
%    \bitbox{2}{0} &
%    \bitbox{2}{$i$} & 
%    \bitbox{3}{$t_{SP}^i$}  &
%    \bitbox{8}{checksum}
%  }\end{rightwordgroup}
%\end{bytefield}

%\begin{bytefield}{19}
%  \begin{rightwordgroup}{response}{
%    \bitbox{7}{preamble} &
%    \bitbox{2}{1} & 
%    \bitbox{2}{$i$} &  
%    \bitbox{8}{checksum}
%  }\end{rightwordgroup}
%\end{bytefield}

%\begin{bytefield}{34}
%  \begin{rightwordgroup}{final}{
%    \bitbox{7}{preamble} &
%    \bitbox{2}{2} & 
%    \bitbox{2}{$i$} &  
%    \bitbox{3}{$t_{SF}^i$} &
%    \bitbox{3}{$t_{RR}^{i,1}$}  &
%    \bitbox{3}{\ldots}  &
%    \bitbox{3}{$t_{RR}^{i,n}$}  &
%    \bitbox{8}{checksum}
%  }\end{rightwordgroup}
%\end{bytefield}
%}

\subsection{Ranging Setup}
Each end cap of SUPERball was fitted with a DWM1000 module located approximately \SI{0.1}{\metre} from the end of the strut.
To simplify the notation, we do not distinguish between end cap positions (ends of the struts) and the positions ranging sensor locations.
In practice, we take this offset into account in the output function of our filter (see Section~\ref{txt:ukf}).

The broadcasting algorithm runs at \SI{15}{\hertz} and packet transmissions are spaced \SI{1}{\milli\second} apart.
This allows for over $20$ modules to range.
After one ranging iteration, each end cap transmits its measurements over WiFi to the ROS network. 
A ROS node then combines measurements from all end caps into a single ROS message at  \SI{10}{\hertz}.

The fixed anchors operate in a similar way to the end caps, but are not connected to a ROS node and can not directly transmit data to the ROS network.
This means that we obtain two measurements (one in each direction) for each pair of modules on the robot, 
but only a single measurement between the fixed anchors and the modules on the robot.

\subsection{Calibration}
\label{txt:calib}
One of the design goals of our state estimation method is quick deployment in new environments without significant manual calibration.
To achieve this, we implemented an automatic calibration procedure to jointly estimate the constellation of fixed modules (anchors, defining an external reference frame) 
and the pairwise sensor offsets ($o_{i,j}$).
Calibration is performed - similar to common motion capture systems - by moving the robot around, while recording the uncorrected distance measurements ($m_{j,i}$).

After recording a dataset, we minimize the reconstruction error $L$ by optimizing over the offsets $\bm{o}$ ($o_{i,j}$ rearranged as a vector), the estimated anchor locations $\bm{N}^{est}$, and the estimated moving module locations $\bm{N}^{float}[1\ldots n_{samples}]$ (i.e. the module on the robot's end caps):
\begin{align}
\resizebox{.91\hsize}{!}{$L\left(i,j,t\right) = \left( \|\bm{N}^{anchor}_i - \bm{N}^{float}_{j}\left[t\right]\|-o_{j,i} - m_{i,j}\left[t\right] \right)^2$ \label{eq:l_single}}\\
\resizebox{.91\hsize}{!}{$L\left(\bm{o},\bm{N}^{anchor},\bm{N}^{float}[1\ldots n_{samples}]\right) = \sum_{i,j,t}\alpha_{j,t}L\left(i,j,t\right)$. \label{eq:l_full}}
\end{align}

The brackets in $\bm{N}^{float}[1\ldots n_{samples}]$ indicate the moving module locations (end cap positions) at a specific timestep. 
For example $\bm{N}^{float}[5]$ contains the estimated end cap positions at timestep 5 in the recorded dataset.
In Eq.~\ref{eq:l_full}, $i$ iterates over anchors, $j$ iterates over moving nodes and $t$ iterates over samples.
The indicator variables $\alpha_{j,t}$ are equal to $1$ when for sample $t$ there are at least $4$ valid measurements to the fixed module for moving module $j$ (i.e. the number of DOFs reduces).

In practice we also add constraints on the bar lengths, which take the same form as Eq.~\ref{eq:l_single} with the offsets set to $0$.
We used BFGS to minimize Eq.~\ref{eq:l_full} with  a dataset containing approximately $400$ timesteps selected randomly from a few minutes of movement of the robot.
Although the algorithm works without prior knowledge, we noticed that providing the relative positions of $3$ fixed nodes ($3$ manual measurements) significantly improves the success rate as there are no guarantees on global convergence.

Once the external offsets (between the anchors and moving nodes) and the module positions are known, we can estimate the offsets between moving nodes in a straightforward way by computing the difference between the estimated internal distances and the uncorrected distance measurements.

%%%%%%%%%%%%%%%%%%%%%%%%%%%%%%%%%%%%%%%%%%%%%%%%%%%%%%%%%%%%%%%%%%%%%%%%%%%%%%%%

\section{Filter Design}
\label{txt:ukf}

Tensegrity systems are nonlinear and exhibit hybrid dynamics due to cable slack conditions and  interactions with the environment that involve collision and friction. This warrants a robust filter design to track the robot's behavior.

The commonly used Extended Kalman Filter (EKF) does not perform well on highly nonlinear systems where first-order approximations offer poor representations of the propagation of uncertainties.
Additionally the EKF requires computation of time-derivatives through system dynamics and output functions which is challenging for a model with complex hybrid dynamics. 

The sigma point UKF does not require derivatives through the system dynamics and is third order accurate when propagating Gaussian Random Variables through nonlinear dynamics \cite{wan2000unscented}. The computational cost of the UKF is comparable to that of the EKF, but for tensegrity systems which commonly have a large range of stiffnesses and a high number of state variables the time-update of the sigma points dominates computational cost. As such we first describe the methods used to reduce computational cost of dynamic simulation, then in the following section we outline the specific implementation of the UKF for the \SB{} prototype.

\subsection{Dynamic Modelling} 

The UKF requires a dynamic model which balances model fidelity and computational efficiency since it requires a large number of simulations to be run in parallel. 
We model a tensegrity system as a spring-mass net and used the following incomplete list of simplifying assumptions:
\begin{itemize}
  \item Only point masses located at each node point
  \item All internal and external forces are applied at nodes
  \item Members exert only linear stiffness and damping
  \item Unilateral forcing in cables
  \item Flat ground at a known height with Coulomb friction
  \item No bar or string collision modelling
\end{itemize}
%This is a common approach for modeling tensegrity systems, and force density approaches for this problem are described in \textcolor{red}{cite}. Below we describe some careful manipulation of the equations within this force density framework which allowed us to run the parallel simulations while leaving computational bandwidth for other requisite operations such as communication and data visualization. 

For a tensegrity with $n$ nodes and $m$ members, the member force densities, 
$\boldsymbol{q}\in\mathbb{R}^{m}$, can be transformed into nodal forces,
$\boldsymbol{F_m}\in\mathbb{R}^{n\times 3}$, by using the current Cartesian nodal positions,
$\boldsymbol{N}\in\mathbb{R}^{n\times 3}$, and the connectivity matrix,
$\boldsymbol{C}\in\mathbb{R}^{m\times n}$, as described in \cite{skelton2009tensegrity}. This operation is described by the equation:
$$
\boldsymbol{F_m} = \boldsymbol{C}^{T} diag(\boldsymbol{q}) \boldsymbol{C} \boldsymbol{N},
$$
where $diag(\cdot)$ represents the creation of a diagonal matrix with the vector argument along its main diagonal.
We first note that $\boldsymbol{C} \boldsymbol{N}$ produces a matrix $\boldsymbol{U}\in\mathbb{R}^{m\times 3}$ where each row corresponds to a vector that points between the $i$th and $j$th nodes spanned by each member. 
Therefore, this first matrix multiplication can be replaced with vector indexing as $\boldsymbol{U}_{k} = \boldsymbol{N}_{i} - \boldsymbol{N}_{j}$,
where we use the notation $\boldsymbol{U}_{k}$ to denote the $k$th row of matrix $\boldsymbol{U}$. If we then compute $\boldsymbol{V}=\boldsymbol{C}\frac{d\boldsymbol{N}}{dt}$ with the same method
%%%% NOTE This was the old sentence, but X_{l} didn't make sense... So I assumed it was suppose to be U_{k}. Is that correct?
%where we use the notation $\boldsymbol{X}_{l}$ to denote the $l$th row of matrix $\boldsymbol{X}$. If we then compute $\boldsymbol{V}=\boldsymbol{C}\frac{d\boldsymbol{N}}{dt}$ with the same method 
as $\boldsymbol{U}$, we obtain a matrix of relative member velocities.
The matrices $\boldsymbol{U}$ and $\boldsymbol{V}$ are used to calculate member lengths as $L_k = |\boldsymbol{U}_k|_2$ and member velocities as $\frac{d}{dt}(L_k) = \frac{\boldsymbol{U}_k(\boldsymbol{V}_k)^T}{L_k}.$

We can then use these values to calculate member force densities, $\boldsymbol{q}$, using Hooke's law and viscous damping as:
$$
\boldsymbol{q}_k = K_k(1 - \frac{L_{0k}}{L_k}) - \frac{c_k}{L_k} \frac{d}{dt}(L_k).
$$
Here $K_k$ and $c_k$ denote the $k$th member's stiffness and damping constants, respectively.
Note that cables require some additional case handling to ensure unilateral forcing.

Scaling each $\boldsymbol{U}_k$ by $\boldsymbol{q}_k$ yields a matrix whose rows correspond to vector forces of the members. 
We denote this matrix as $\boldsymbol{U}^q\in\mathbb{R}^{m\times 3}$, and we note that $\boldsymbol{U}^q = diag(\boldsymbol{q}) \boldsymbol{C} \boldsymbol{N}$.
Thus this matrix of member forces can be easily applied to the nodes using:
$$
\boldsymbol{F_m} = \boldsymbol{C}^{T} \boldsymbol{U}^q.
$$
We now have a method for computing nodal forces exerted by the members and need only compute ground interaction forces, which we will denote as $\boldsymbol{F}_g$.
We computed ground interaction forces using the numerical approach in~\cite{yamane2006stable}. The nodal accelerations can then be written as:
$$
\frac{d^2\boldsymbol{N}}{dt^2} = \boldsymbol{M}^{-1}(\boldsymbol{F}_m+ \boldsymbol{F}_g) -  \boldsymbol{G},
$$
where $\boldsymbol{M}\in\mathbb{R}^{n\times n}$ is a diagonal matrix whose diagonal entries are the masses of each node and $\boldsymbol{G}\in\mathbb{R}^{n\times 3}$ is matrix with identical rows equal to the vector acceleration due to gravity. 
It is then straightforward to simulate this second order ODE using traditional numerical methods. 

Note also that it is possible to propagate many parallel simulations efficiently by concatenating multiple $\boldsymbol{N}$ matrices column wise to produce  $\boldsymbol{N}_{\parallel}\in\mathbb{R}^{n\times 3l}$ for $l$ parallel simulations.
The resultant vectorization of many of the operations yields significant gains in computational speed with some careful handling of matrix dimensions.   

\subsection{UKF Implementation}

We implement a traditional UKF as outlined in \cite{wan2000unscented} with additive Gaussian noise for state variables and measurements.

Several parameters are defined for tuning the behavior of the UKF, namely $\alpha$, $\beta$ and $\kappa$, where $\alpha$ determines the spread of the sigma points generated by the unscented transformation, $\beta$ is used to incorporate prior knowledge of distribution, and $\kappa$ is a secondary scaling parameter.
We hand tuned these parameters to the values $\alpha = 0.0139$, $\beta = 2$ for Gaussian distributions and $\kappa = 0$ and found this to yield an adequately stable filter.

We define our state variables as $\boldsymbol{N}$ and $\frac{d\boldsymbol{N}}{dt}$ stacked in a vector $\boldsymbol{y}\in\mathbb{R}^{L}$ where $L = 6n$ is the number of state variables. 
We assume independent state noise with variance $\lambda_y = 0.4$ thus with covariance $\boldsymbol{R} = \lambda_y\bm{I}_L$.

%For measurements we take the minimum angle between each bar vector and the z-axis, $\theta\in\mathbb{R}^{b}$ where $b$ is the number of bar angles available at the given time step and all ranging measures, $\boldsymbol{r}\in\mathbb{R}^{a}$, where $a$ is the number of ranging measures available at a given time step.
For measurements we take estimated orientation data from our IMUs using a gradient descent AHRS algorithm based on~\cite{madgwick2011estimation}, $\theta\in\mathbb{R}^{b}$ where $b$ is the number of bar angles available at the given time step and all ranging measures, $\boldsymbol{r}\in\mathbb{R}^{a}$, where $a$ is the number of ranging measures available at a given time step.
We again assume independent noise with $\lambda_\theta = 0.1$ and $\lambda_r = 0.029$ the measurement covariance matrix is then defined as:
$$
\boldsymbol{Q} =  \left[ \begin{array}{ccc} \lambda_\theta\bm{I}_b & \boldsymbol{0} \\
                         \boldsymbol{0}        & \lambda_r\bm{I}_a  \end{array} \right].
$$
These user defined variables are then used within the framework of our UKF to forward propagate both the current expected value of the state as well as its covariance. 
Fig.~\ref{fig:UKFflowChart} shows an overview of our complete state estimation setup.

\begin{figure}[tpbh]
 \centering
  \includegraphics[width=0.85\linewidth]{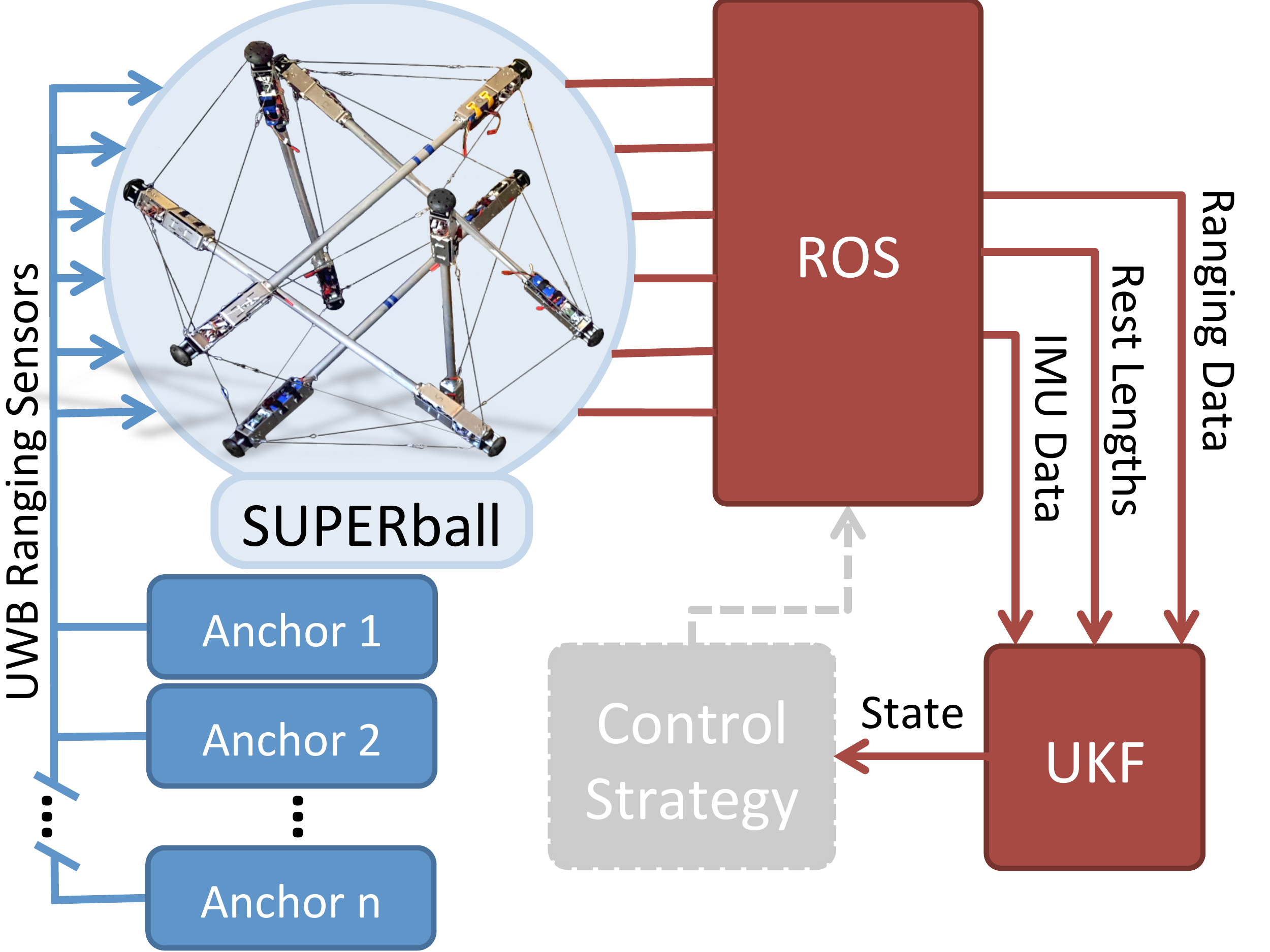}
 \caption{Block diagram of data flow within the system. Red signals are passed as ROS messages and blue signals are passed using the ranging modules. Note that each rod contains two ranging sensors located at each end of the rod. The gray control strategy block represents a to-be-designed state-feedback control strategy.}
\label{fig:UKFflowChart}
 \end{figure}

\section{Filter Evaluation}
\subsection{Experimental Setup}

\begin{figure}[tpbh]
 \centering
  \includegraphics[width=\linewidth]{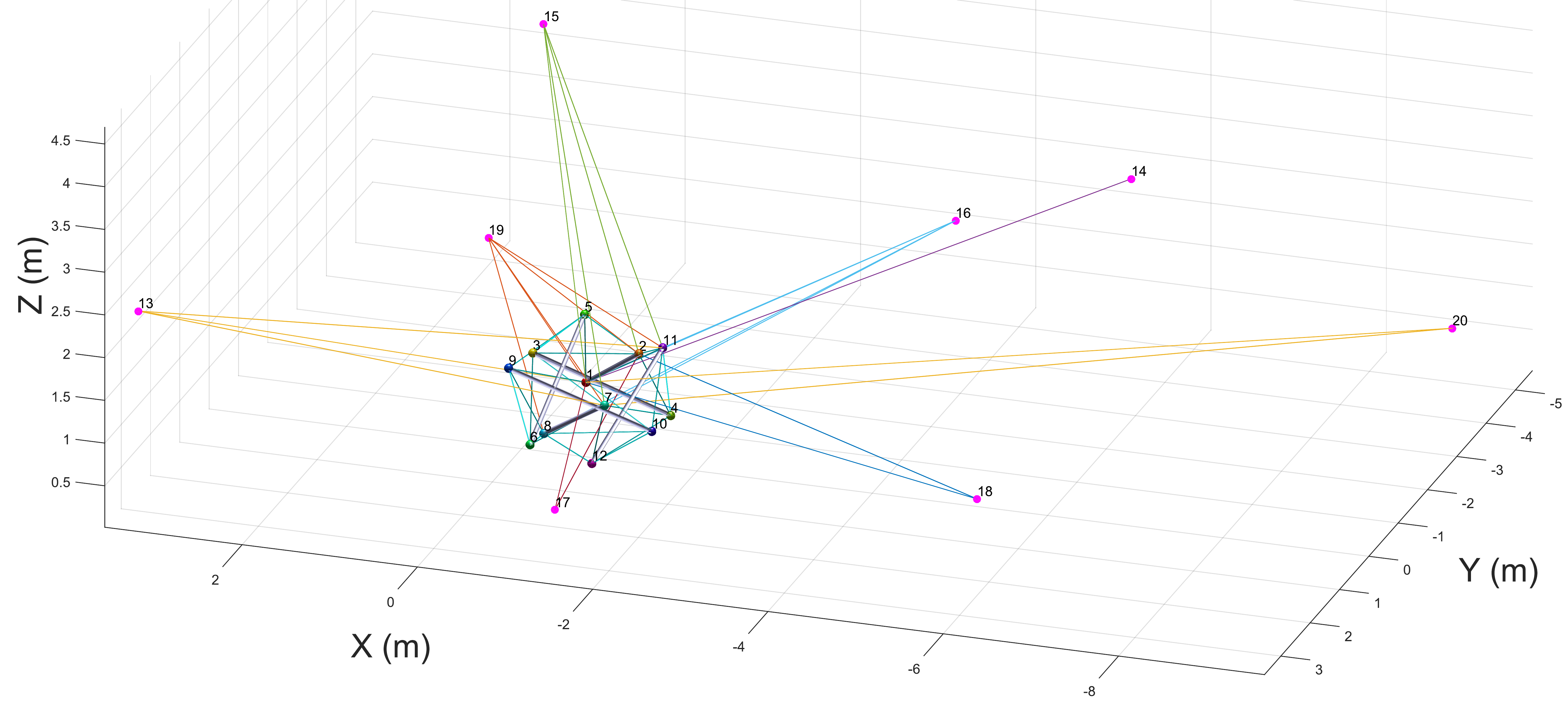}
 \caption{Visualization of the UKF output. \SB{} sits in the middle of the plot surrounded by 8 ranging base stations. Lines between the robot and the base stations indicate valid ranging measures during this timestep.}
\label{fig:SUPERballMATLAB}
 \end{figure}
 
To evaluate the performance of the UKF, we used the eight "fixed anchor" ranging base stations calibrated as detailed in Section~\ref{txt:calib}.
Each end cap of \SB{} was then able to get a distance measurement to each base station.
This information was sent over ROS along with IMU data (yaw,pitch,roll) and cable rest lengths to the UKF.
The base stations were placed in a pattern to cover an area of approximately \SI{91}{\meter^2}. 
Each base station's relative location to each other may be seen in Fig.~\ref{fig:SUPERballMATLAB}.
\SB{} and the base stations were then used to show the UKF tracking a local trajectory of end caps and a global trajectory of the robotic system.
In each of these experiments, the UKF was allowed time to settle from initial conditions upon starting the filter.
This ensured that any erroneous states due to poor initial conditioning did not affect the filter's overall performance. 

\subsection{Local Trajectory Tracking}
In order to track a local trajectory, \SB{} remained stationary while two of its actuators tracked phase shifted stepwise sinusoidal patterns.
During the period of actuation, two end cap trajectories were tracked on \SB{} and compared to the trajectory outputs of the UKF.
One end cap was directly connected to an actuated cable (end cap 2), while the other end cap had no actuated cables affixed to it (end cap 1).
To obtain a ground truth for the position trajectory, a camera that measured the position of each end cap was positioned next to the robot.
Both end caps started at the same relative height and the majority of movement of both fell within the plane parallel to the camera.
Fig.~\ref{fig:smalldisplacement} shows the measured and UKF global positions of the two end caps through time.

\begin{figure}[tpbh]
  \centering
  \includegraphics[width=1\linewidth]{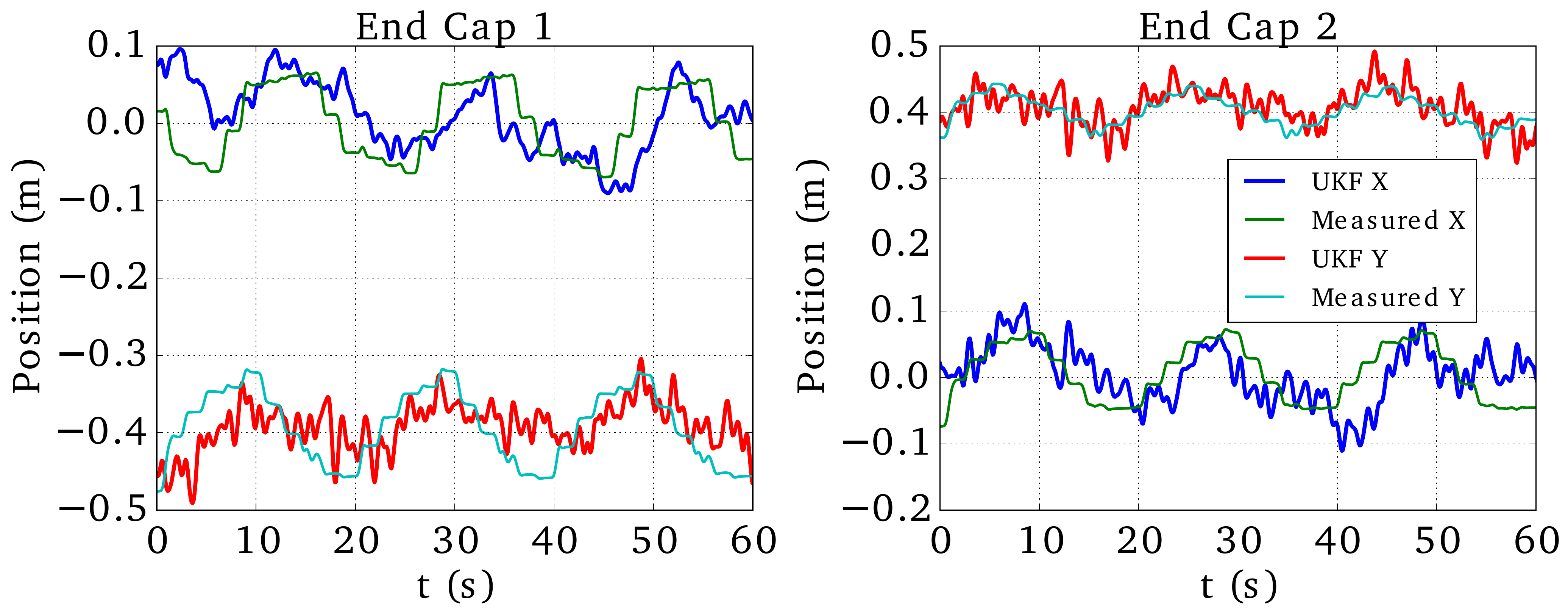}
  \caption{Position plotted through time for both end cap 1 and end cap 2. The thin line represents the position output measured by the camera tracking system, and the bold line represents the position output from the UKF filter. As expected, there is a time domain lag between the measured and estimated positions.}
  \label{fig:smalldisplacement}
\end{figure}

% For this experiment, the cables between end caps 1 and 11 and end caps 12 and 8 were actuated.
% The UKF is able to track the end cap movements quite well with some displacement error in the Y position for end cap 1.
% Upon further inspection of the input data to the UKF, there was a high packet loss between end cap 1 and the base stations. 
% This coupled with a mismatched base model, might be the cause for this error.

\subsection{Global Trajectory Tracking}
For global trajectory tracking, \SB{} was actuated to induce a transition from one base triangle rolling through to another base triangle as presented in \cite{sabelhaus2015system}.
%The state of \SB{} was tracked using the UKF.
Ground truth for this experiment was ascertained by marking and measuring the positions of each base triangle's end caps before and after a face transition.
%Fig.~\ref{fig:3roll_perspective} shows a 3D plot of the UKF generated states for the beginning and end of the experiment.
%Each colored triangle represents a base triangle and the robot implements two full transitions starting from the red triangle and ending on the blue.
We evaluate 4 settings of the state estimator. \emph{Full}: The state estimator as described in Section~\ref{txt:ukf} with all IMU and ranging sensors. \emph{no IMU}: Only the ranging sensors are enabled. \emph{full w. cst. offset}: Same as \emph{full}, but the offsets $\bm{o}$ are set to a constant instead of optimized individually. \emph{4 base station ranging sensors}: 50\% of the base station ranging sensors are disabled. 
The results of this experiment are presented in Fig.~\ref{fig:3roll_triangles}~and~\ref{fig:3roll_xz_position}.
\begin{figure}[tpbh]
 \centering
  \includegraphics[width=\linewidth]{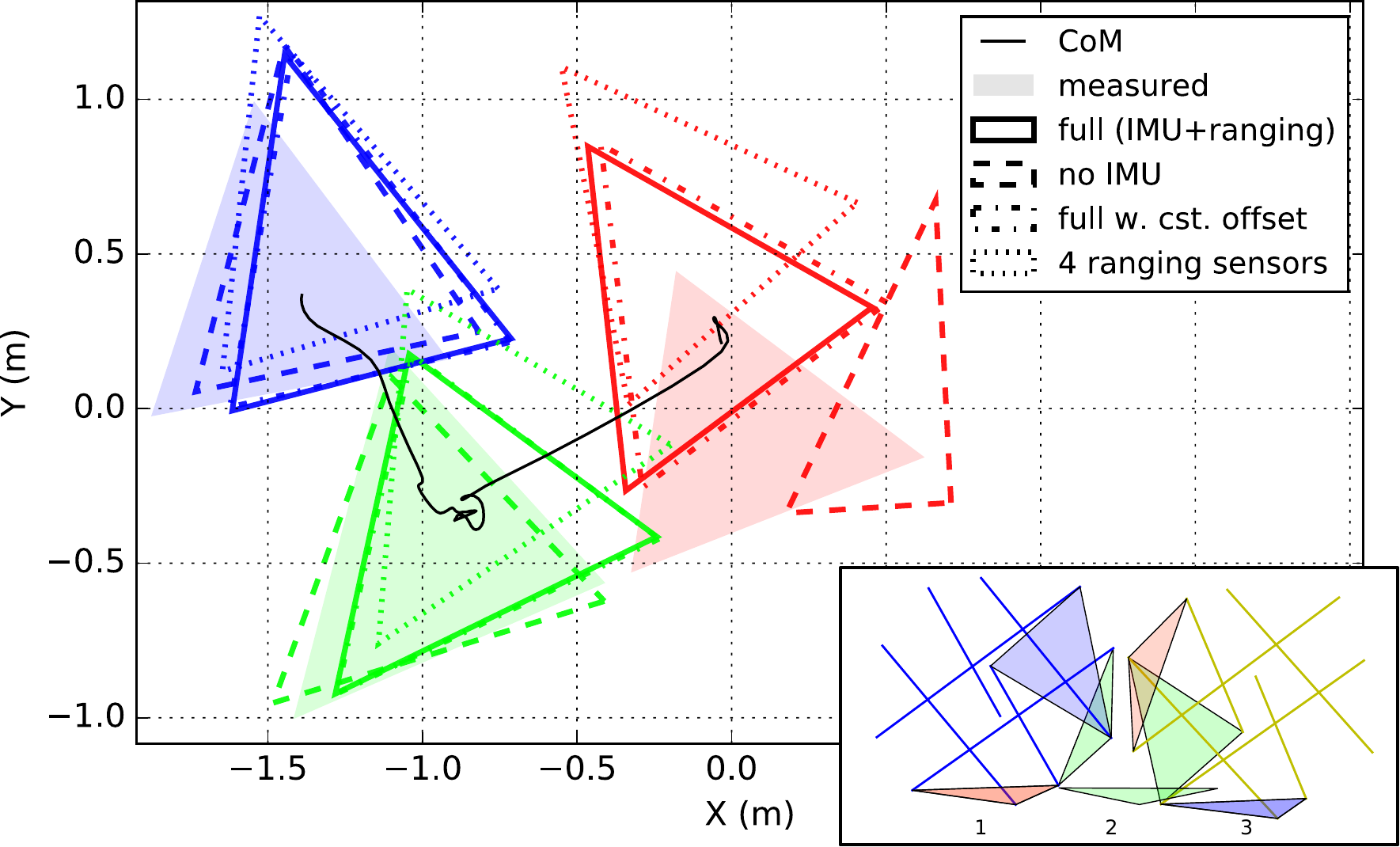}
\caption{Top down view of the triangular faces to which the robot transitions during the global trajectory tracking experiment for various setting of the state estimator. The small inset illustrates the movement of the robot. The line shows the estimated center of mass (CoM) using the \emph{full} settings.
Finding the initial position (origin) is hard for all settings, and without the IMUs the estimator does not find the correct initial face.
After a first roll, tracking becomes more accurate. The offsets $\bm{o}$ have a minimal impact, which indicates that our calibration routine is sufficiently accurate.   }
\label{fig:3roll_triangles}
 \end{figure}
\begin{figure}[tpbh]
  \centering
  \includegraphics[width=\linewidth]{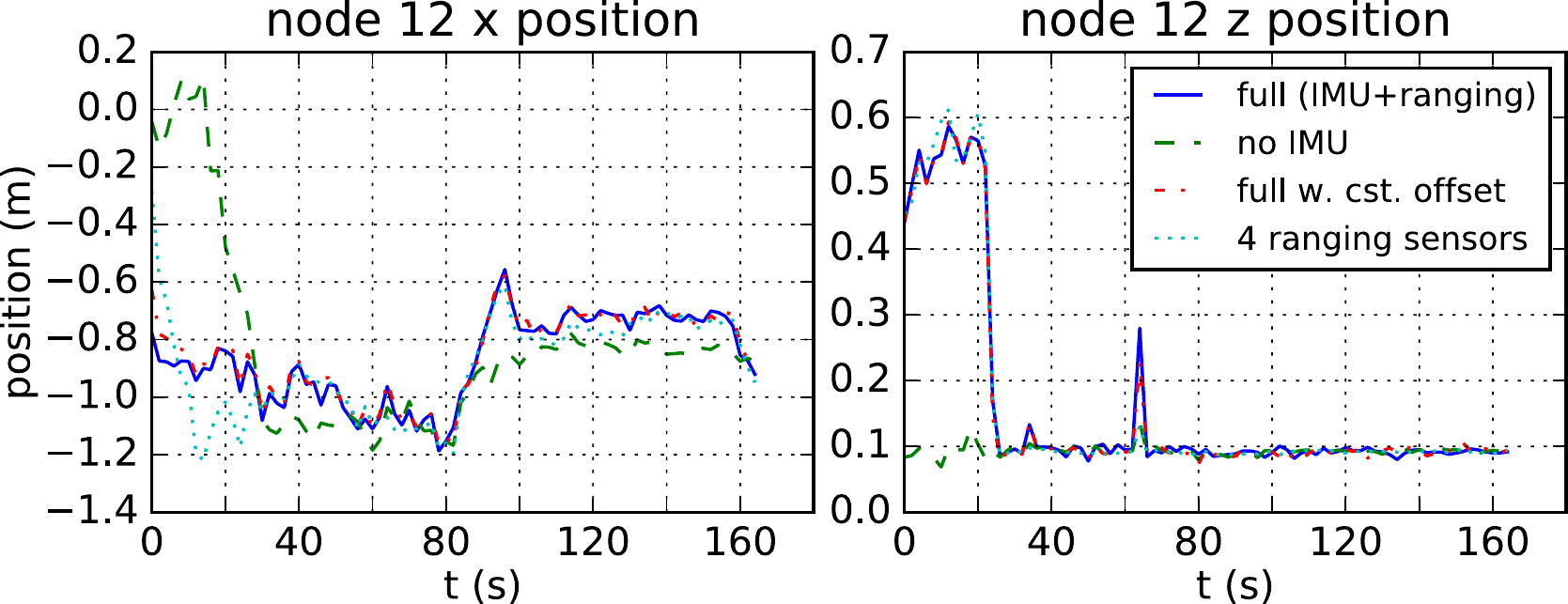}
\caption{X and Y position of end cap 12 as a function of time for the various estimator settings. The end cap was initially off the ground and touches the ground after the first roll. This is not tracked correctly when the IMUs are disabled. The system works as expected when 4 base stations ranging sensors are disabled, but with slower convergence and more noise on the robot's position. Around \SI{60}{s} there's a spurious IMU value from which  the state estimator recovers. 
}
  \label{fig:3roll_xz_position}
\end{figure}

%After settling was allowed from initial conditions, the filter tracked the position of each base triangle quite well with an maximum error of \SI{0.2}{\meter} at any point in time.
%Since the robot was static after each base triangle transition, the center of mass of the robot may be analogous to the center of each base triangle.
%Fig.~\ref{fig:3roll_top_view} presents a snap shot of each base triangle overlaid onto the measured ground truth triangles.
%The black dots represent the center of mass of the estimated states obtained from the UKF.

\section{Conclusion}
We have introduced a state estimation approach for tensegrity robots based on ultra wideband ranging sensors, inertial measurements, and actuator states.
An unscented Kalman filter was used to combine these sensor and state observations.
While this is a fairly common approach to state estimation, our algorithm is robust to measurement noise and significant amounts of missing data.
To verify these statements, we evaluated our method on two tasks: rolling and stationary deformations of \SB{}.

Most of the current control approaches for tensegrity robots rely on position tracking for either motion planning or performance evaluation.
Hence, our main contribution to the field of tensegrity robotics is that this work will finally allow various proposed control algorithms for tensegrity robots to be transferred from the simulation domain to hardware.

We believe that - in this context - our setup is a viable alternative to more established external solutions, such as motion capture systems.
In particular, our approach is low-cost, self-calibrating and only relies on autonomous anchors.
These last two qualities are particularly attractive for future space missions.
We do not yet achieve the accuracy provided by commercial motion capture systems.
However, this is not a priority at this point as we mainly are focused on large displacements by rolling of \SB{}.

The logical next step is to test our state estimator when SUPERball is moving around in a larger space. 
The RoverScape at NASA Ames - a football field sized outdoor rover testbed - is the ideal candidate test area for this.

A related future goal is to add support for an incremental number of ranging anchors. 
This will allow \SB{} to explore uncharted terrain by dropping beacons when the uncertainty of its current position increases.

% Measuring the true lengths of compliant tension elements in a tensegrity system would benefit the accuracy of state estimation. 
% Unfortunately, this is non-trivial from an engineering perspective though some new soft sensors may prove useful in future work \cite{vogt2013design}.  


%%%%%%%%%%%%%%%%%%%%%%%%%%%%%%%%%%%%%%%%%%%%%%%%%%%%%%%%%%%%%%%%%%%%%%%%%%%%%%%%
{\small
\section*{Acknowledgments}
We appreciate the support, ideas, and feedback from members of the Dynamic Tensegrity Robotics Lab.  
We are also grateful to Terry Fong and the NASA Ames Intelligent Robotics Group.  
}

%%%%%%%%%%%%%%%%%%%%%%%%%%%%%%%%%%%%%%%%%%%%%%%%%%%%%%%%%%%%%%%%%%%%%%%%%%%%%%%%

\bibliographystyle{IEEEtran}
\balance
\bibliography{IEEEabrv,ICRA_2016_SUPERball.bib}

\end{document}